\pdfoutput=1
\documentclass[11pt]{article}
\usepackage{amsmath}
\usepackage{amssymb}
\usepackage{booktabs}
\usepackage{float}

\usepackage[preprint]{acl}

\usepackage{times}
\usepackage{latexsym}

\usepackage[T1]{fontenc}

\usepackage[utf8]{inputenc}

\usepackage{microtype}

\usepackage{inconsolata}

\usepackage{graphicx}

\title{Change the Number, Change the Answer: \\
Assessing Rating Scale Sensitivity in LLM-Based Text Scoring}

\author{Wonjeong Jo \\
  UC Davis \\
  \texttt{wjo@ucdavis.edu} \And
  Gyuwan Kim \\
  UC Santa Barbara \\
  \texttt{gyuwankim@ucsb.edu} \And
  Ming Shan Hee \\
  MBZUAI \\
  \texttt{mingshan.hee@mbzuai.ac.ae}}

\begin{document}
\maketitle
\begin{abstract}
Large language models (LLMs) are increasingly used in computational social science to assign scalar scores to social and psychological constructs. However, studies use different rating scales, and whether their scores remain comparable across scale designs is understudied. To address this gap, we define and systematically investigate \textit{scale sensitivity}, the variation in LLM-generated scores caused by rating scale design. We distinguish \textit{numerical sensitivity}, changes in normalized scores, from \textit{semantic sensitivity}, changes in score interpretation. To assess both, our experimental design crosses 2 formats (continuous vs.\ ordinal), 4 ranges (3-, 5-, 11-, and 21-point), and 2 signs (signed vs.\ unsigned) across 3 tasks (political leaning, sentiment, and toxicity), 2 prompting strategies (zero-shot and rubric-based), and 5 LLMs. We find that changing one design element shifts normalized scores by about 0.10 on average and changes the category of roughly 20\% of texts. Sensitivity is greatest for toxicity, the only unipolar construct examined. Also, rubric-based prompting does not reduce sensitivity compared with zero-shot prompting. We provide practical guidance for designing more reliable LLM-based text scoring and highlight the need for robustness checks across alternative scales.

\end{abstract}

\section{Introduction}
\label{Introduction}

Large language models (LLMs) are increasingly used for text analysis at scale. Compared to human raters or traditional supervised models, LLMs provide strong cross-domain versatility with reduced costs and time. In computational social science, scholars use LLMs for various tasks, such as political leaning identification \cite{gilardi2023chatgpt, heseltine2024large, tornberg2025large}, hateful or toxic content detection \cite{hee2025demystifying, JoWojcieszak2024HarmfulYouTube}, sentiment analysis \cite{rathje2024gpt, ornstein2025train}, and topic or framing analysis \cite{kim2024accuracy,
wang2023prompting, kuang2025towards}.

In text analysis, many constructs are better understood when measured along a continuum rather than as discrete categories \cite{licht2025measuring}. For example, concepts such as ideological extremity, toxicity severity, and moral emotions vary in degree rather than kind. To incorporate these nuances and gradations, a growing number of researchers prompt LLMs to generate scalar scores on numerical scales, treating model outputs as continuous or ordinal measures of underlying latent constructs \cite{le2025positioning, koh2024can, li2024hot}.

\begin{figure}
    \centering
    \includegraphics[width=\columnwidth]{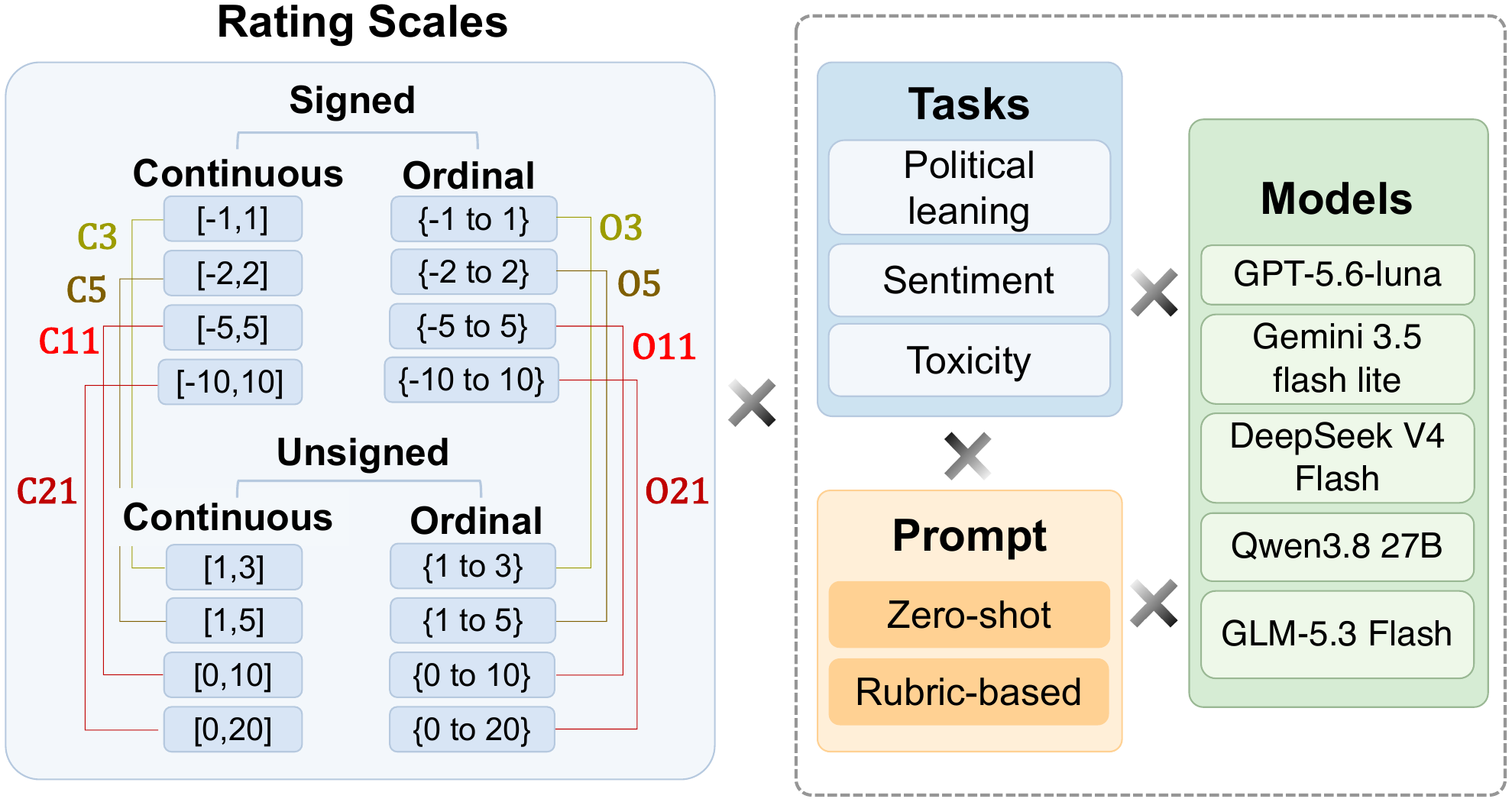}
    \caption{\footnotesize Experimental design for evaluating scale sensitivity. 
    }
    \label{fig:overview}
\end{figure}

Yet, how consistent their outputs are across different rating scales or how sensitive those outputs are to changes in scale design has received limited academic attention. Existing studies have relied on different prompts and rating scale designs that vary in both scale format and range, even when measuring the same underlying construct. For example, some studies evaluate toxicity using a continuous scale between 0 and 1 \cite{li2024hot, koh2024can}, whereas other studies adopt Likert-type ordinal scales ranging from 1 to 5 \cite{movva2024annotation}. As such, systematic examination of how rating scale design influences LLM-generated scores is essential to establish the reliability, validity, and comparability of LLM-based measurement. 

To address this gap, we define and introduce \textbf{scale sensitivity} as the degree to which LLM-generated scalar outputs vary in response to changes in rating scale design. We systematically investigate three design elements of the rating scale: \textit{format} (continuous vs. ordinal), \textit{range} (3-, 5-, 11-, and 21-point scales), and \textit{sign} (signed scales centered at 0 vs. unsigned scales containing only non-negative values). We assess scale sensitivity across tasks with different conceptual foundations (political leaning, toxicity, and sentiment), prompt designs (zero-shot vs. rubric-based), and models.

We further distinguish between two forms of scale sensitivity: \textbf{numerical sensitivity} and \textbf{semantic sensitivity}. Numerical sensitivity captures whether a model assigns different normalized scores to the same item when the rating scale changes, whereas semantic sensitivity captures whether those changes switch the substantive interpretation of the output. They reflect different consequences of a change in the rating scale, that is, how far a score moves and whether its interpretation changes. For example, consider a normalized toxicity scale from 0 to 1 that is divided into three equal categories. If a text's score moves from 0.30 to 0.36, the numerical change is small, but the text moves from the “not toxic” category to the “moderately toxic” category. In contrast, if a text's score moves from 0.70 to 0.95, the numerical change is much larger, but the text remains in the “strongly toxic” category. 

Using the experimental design shown in Figure~\ref{fig:overview}, we find empirical evidence of both numerical and semantic scale sensitivity. Specifically, changing one rating scale design element shifts normalized scores by about 0.10 and changes the category of about 20\% of texts. The toxicity scoring task shows the greatest sensitivity, particularly to scale sign. Compared with zero-shot prompting, rubric-based prompting does not reduce sensitivity. We additionally test whether different levels of reasoning effort affect scale sensitivity and show that greater reasoning effort can reduce scale sensitivity in open models. 

Our paper makes three key contributions.
\begin{itemize}
    \item We conceptualize and empirically characterize \textit{scale sensitivity} as an underexamined source of variation in LLM-based measurement.
    \item We demonstrate that scale sensitivity has direct implications for the comparability, replicability, and validity of LLM-scored measurements.
    \item We provide practical guidance for more reliable LLM-based text scoring. We recommend robustness checks across alternative rating scales, choosing scales that match the structure of the construct, and assessing agreement with suitable human labels, when available, as evidence of validity.
\end{itemize}

\section{Literature Review}
\label{sec:related}

\subsection{Rating Scale Design in Measurement}
Survey and psychometric research demonstrates that rating scale design affects how humans rate the same construct. We review previous research for the three design elements we manipulate: format, range, and sign.

\paragraph{Format.} Social scientists typically measure scalar constructs with two broad families of rating scales. For one, attitudes, emotions, and perceptions have traditionally been measured with ordinal Likert-type scales (e.g., 1--5 or 1--7) \citep{likert2017method, anastasiadou2011reliability} \footnote{Although Likert scales are often treated as interval, we regard them as ordinal because the distances between scale points are not guaranteed to be equal \citep{WuLeung2017LikertInterval}.}. For another, computational tools increasingly assign continuous scores, including VADER sentiment scores ranging from $-1$ to $1$ \citep{HuttoGilbert2014VADER} and Perspective API toxicity scores ranging from $0$ to $1$ \citep{PerspectiveAPIAttributes}. Compared with ordinal scales, continuous scales preserve more information and allow finer-grained judgments \citep{Treiblmaier2011ContinuousScales}. However, studies comparing the two formats have not found that either one consistently yields higher reliability or validity \citep{Treiblmaier2011ContinuousScales, svensson2000comparison}. Regardless of which format is preferable, the two formats tend to produce different ratings of the same stimuli \citep{lawless1986comparison}.

\paragraph{Range.} Rating scales differ in range, that is, the number of points they offer (e.g., 3, 5, 11, or 21 points). Changing the number of points changes the measurement properties of a scale, but findings are inconclusive on whether more points are better. Some studies suggest that adding points can improve distributional properties and internal consistency \citep{leung2011comparison, xu2018effects, chang1994psychometric}, which may in turn improve validity and reliability \citep{kusmaryono2022number}. Others find that overly fine-grained scales may reduce consistency and reliability in certain contexts \citep{cai2016study, chang1994psychometric}.

\paragraph{Sign.} A rating scale with the same number of points can be signed or unsigned. A signed scale includes negative values and is centered on 0 (e.g., $-1$ to $1$), whereas an unsigned scale uses only non-negative values (e.g., $1$ to $3$). The sign of the scale changes how respondents interpret its endpoints. For example in an empirical study, when the same verbal endpoints were labeled $-5$ to $+5$ instead of $0$ to $10$, human respondents selected the lower half of the scale less often. Negative numbers led them to interpret the low endpoint as the presence of an opposite attribute rather than the absence of the attribute \citep{schwarz1991rating}. Similarly, labeling a 7-point scale $-3$ to $+3$ instead of $1$ to $7$ shifted responses toward the positive end \citep{tourangeau2007color}. In short, scale sign can influence both the distribution of ratings and the meaning respondents attach to scale points.

\subsection{Rating Scale Design in LLM Prompts}
It remains unclear whether rating scale design influences LLM-generated scores as it does human ratings. Thus, this paper looks into how consistent LLM-generated scores are across rating scale designs. Given that LLM outputs are sensitive to changes in prompt formatting, ordering, and instructions \citep{sclar2023quantifying, he2024does, guan2025order, zhuo2024prosa}, we hold the prompt fixed within each prompting strategy and change only the rating scale it specifies so that differences in scores across rating scale conditions can be attributed to scale design. We quantify these differences as numerical and semantic sensitivity. In this context, we ask: \textbf{(RQ1)} To what extent do changes in rating scale format, range, and sign yield (a) numerical and (b) semantic scale sensitivity in LLM-generated outputs?

\subsection{Variance Across Task, Prompting Strategy, and Model}

We further examine whether three factors moderate scale sensitivity, namely the task, the prompting strategy, and the model.

\paragraph{Task.} We measure political leaning, sentiment, and toxicity, three constructs widely explored in computational social science \citep{le2025positioning, robertson2023negativity, rathje2021out, yu2024partisanship}. Political leaning, such as positioning actors or texts along a left-right spectrum, is a \textit{social construct}, grounded in political parties, agendas, elites, and symbolic meanings that are socially agreed upon \cite{bergan2012partisan, boudreau2014informing}. Sentiment is a \textit{psychological construct} that reflects individuals’ affective states such as positivity or negativity \cite{evans2002emotion}. Toxicity is a \textit{socio-psychological} construct whose judgment depends on both social norms and expressions of anger or hostility \citep{sheth2022defining, Pandita2024ThreeRoots}. We note that political leaning and sentiment are bipolar constructs that range between two opposing poles (left vs.\ right and negative vs.\ positive), whereas toxicity is a unipolar construct that ranges from the absence of toxicity to its strong presence. Considering these conceptual differences across tasks, we ask: \textbf{(RQ2)} To what extent does the task moderate (a) numerical and (b) semantic scale sensitivity? 

\paragraph{Prompting Strategy.} The prompting strategy may also affect how stable LLM-generated scores are across rating scales. We compare two strategies that do not introduce additional confounding factors and differ in how much guidance they provide: zero-shot and rubric-based prompting strategies. Zero-shot prompting is commonly used in LLM-based text scoring \citep{gilardi2023chatgpt, tornberg2025large, le2025positioning} and requests a direct rating without explicit evaluation criteria. Rubric-based prompting, modeled on educational assessment, provides detailed coding guidelines for predefined dimensions of a construct. It asks the model to score each dimension and use these assessments to determine a final rating. Rubrics have been shown to improve consistency in LLM-based scoring \citep{hashemi2024llm, eltanbouly2025trates, sharma2025researchrubrics, li2026rubricem}, but whether they also stabilize scores across rating scales is unknown. If scale sensitivity stems from underspecified evaluation criteria, explicit rubrics should reduce it. If it stems from how models map judgments onto numerical scales, it should persist regardless of prompting strategy. As such, we ask: \textbf{(RQ3)} To what extent does the prompting strategy (zero-shot vs.\ rubric-based) moderate (a) numerical and (b) semantic scale sensitivity?

\paragraph{Model.} Model selection has become a methodological choice in computational social science. Each model differs in training data, optimization objectives, and alignment procedures, and most offer limited transparency into how they represent semantic judgments internally \citep{kristensen2023chatbots}. Considering these differences, we ask: \textbf{(RQ4)} To what extent does the choice of LLM moderate (a) numerical and (b) semantic scale sensitivity?

Additionally, we test which measurement design choices most strongly affect the sensitivity of LLM-generated scores by asking: \textbf{(RQ5)} How much of the variation in (a) numerical and (b) semantic scale sensitivity is explained by scale design, task, prompting strategy, and model?

\section{Experimental Design}
\label{experiment design}

\subsection{Dataset}
We use three publicly available datasets, one for each task (see Table~\ref{tab:datasets}). Each dataset contains texts labeled by human annotators. However, we do not use these human labels in our main analysis because our focus is on the consistency of AI-generated outputs across rating scales rather than agreement between humans and models. Instead, we incorporate the human labels for supplementary analyses reported in Section~\ref{sec:add} alongside our main findings. We describe each dataset in detail below.

\subsubsection{Political Leaning}
For political leaning, we use the MBIC (Media Bias Including Characteristics; \citealp{Spinde2021MBIC}).\footnote{\url{https://www.kaggle.com/datasets/timospinde/mbic-a-media-bias-annotation-dataset/data}, CC BY-NC-SA 4.0} MBIC contains 1,700 sentences selected from approximately 1,000 articles published by eight U.S. news outlets between January 2019 and June 2020. The outlets span the political spectrum and were categorized as left-, center-, or right-leaning based on ratings from AllSides and Ad Fontes Media. Each sentence was evaluated by ten U.S.-based crowdworkers for linguistic bias and opinionatedness. We use all 1,700 sentences for our analysis.

\subsubsection{Sentiment}
For sentiment, we use an annotated dataset published by VADER (Valence Aware Dictionary and sEntiment Reasoner) \footnote{\url{https://github.com/cjhutto/vaderSentiment}, MIT License}, a lexicon- and rule-based sentiment analysis tool \cite{HuttoGilbert2014VADER}. Specifically, we used the \textit{movie review} subset, which includes 10,605 Rotten Tomatoes reviews labeled by 20 independent human raters on a scale from $-4$ (extremely negative) to $4$ (extremely positive). From this dataset, we sample 5,000 reviews for the test set.

\subsubsection{Toxicity}
For toxicity, we use the dataset from the \textit{Jigsaw Unintended Bias in Toxicity Classification} competition\footnote{\begingroup\def\UrlBreaks{\do\/\do-\do\.\do\_}\url{https://www.kaggle.com/competitions/jigsaw-unintended-bias-in-toxicity-classification}\endgroup, CC0} \cite{jigsaw-unintended-bias-in-toxicity-classification}. The training dataset contains 1,804,874 public comments posted between 2015 and 2017 on Civil Comments, an online commenting platform used by around 50 English-language news sites. Up to ten human annotators rated each comment as "very toxic," "toxic," "hard to say," or "not toxic." The dataset’s aggregated toxicity score ranges from 0 to 1 and represents the proportion of annotators who rated a comment as toxic or very toxic. From the training dataset, we sample 5,000 comments to form the test set used in our study.

\begin{table}[h]
\centering
\footnotesize
\setlength{\tabcolsep}{2pt}
\renewcommand{\arraystretch}{1.15}

\begin{tabular*}{\linewidth}{@{\extracolsep{\fill}}p{0.32\linewidth}p{0.43\linewidth}r@{}}
\hline
\textbf{Construct} & \textbf{Data Source} & \textbf{Sampled $N$} \\
\hline
Political leaning & MBIC (Media Bias Including Characteristics) \cite{Spinde2021MBIC} & 1,700 \\
Sentiment & VADER movie reviews \cite{HuttoGilbert2014VADER} & 5,000 \\
Toxicity & Jigsaw Unintended Bias in Toxicity Classification \cite{jigsaw-unintended-bias-in-toxicity-classification} & 5,000 \\
\hline
\end{tabular*}

\caption{Data sources and sample sizes used in the study. Sample size refers to the number of items extracted from each original dataset and analyzed in our study.}
\label{tab:datasets}
\end{table}

\subsection{Rating Scale Conditions}

We vary three rating scale design elements. Response format is continuous (any real number within the range) or ordinal (integers only). Scale range is 3-, 5-, 11-, or 21-point. Scale sign is signed (centered on 0) or unsigned (non-negative values only). Each unique combination of these elements defines a rating scale condition, yielding 16 conditions in total (see Table~\ref{tab:scale-conditions}).

\begin{table}[h]
\centering
\scriptsize
\setlength{\tabcolsep}{3pt}
\renewcommand{\arraystretch}{1.05}

\begin{tabular*}{\columnwidth}{@{\extracolsep{\fill}}llcc@{}}
\hline
\textbf{Format} & \textbf{Sign} & \textbf{Scale bounds} & \textbf{Range} \\
\hline

Continuous & Signed & $[-1, +1]$ & 3 \\
           &        & $[-2, +2]$ & 5 \\
           &        & $[-5, +5]$ & 11 \\
           &        & $[-10, +10]$ & 21 \\
\cline{2-4}
           & Unsigned & $[1, 3]$ & 3 \\
           &          & $[1, 5]$ & 5 \\
           &          & $[0, 10]$ & 11 \\
           &          & $[0, 20]$ & 21 \\

\hline

Ordinal & Signed & $-1$ to $+1$ & 3 \\
        &        & $-2$ to $+2$ & 5 \\
        &        & $-5$ to $+5$ & 11 \\
        &        & $-10$ to $+10$ & 21 \\
\cline{2-4}
        & Unsigned & 1 to 3 & 3 \\
        &          & 1 to 5 & 5 \\
        &          & 0 to 10 & 11 \\
        &          & 0 to 20 & 21 \\

\hline
\end{tabular*}

\caption{The 16 rating scale conditions tested in the study.}
\label{tab:scale-conditions}
\end{table}

\subsection{Models} 
We compare scores on these rating scale conditions across 
\texttt{GPT 5.6 Luna}~\cite{openai2026gpt56luna},
\texttt{Gemini 3.5 Flash Lite}~\cite{google2026gemini35flashlite},
\texttt{DeepSeek V4 Flash}~\cite{deepseekai2026deepseekv4},
\texttt{Qwen3.8 27B}~\cite{qwen2026qwen38_27b} and
\texttt{GLM 5.3 Flash}~\cite{zai2026glm53flash}.

The three open models (DeepSeek V4 Flash, Qwen3.8 27B, and GLM 5.3 Flash) were self-hosted on a single fixed endpoint with the sampling settings recommended on their Hugging Face pages. The two proprietary models (GPT 5.6 Luna and Gemini 3.5 Flash Lite) were accessed through each provider's own API with the default sampling settings. We ran every model at its lowest reasoning level, \texttt{low} for the open models and GPT 5.6 Luna and \texttt{minimal} for Gemini 3.5 Flash Lite. Appendix~\ref{app:llm_parameters} provides details about the model parameters. 

\subsection{Prompts}
\label{sec:prompts}

We compare \textit{zero-shot} and \textit{rubric-based} prompting. The two prompts share the same task description, scale instruction, scoring instructions, and input text. The rubric-based prompt adds the rubric, instructions for scoring each dimension, and an output field for each dimension. Details about the prompts are presented in Appendix~\ref{app:prompts}.

\subsubsection{Zero-Shot Prompt}

The zero-shot prompt asks the model to evaluate the target construct without detailed coding guidelines or examples. For example, it asks the model to “\textit{Examine and evaluate the overall political leaning expressed in the given U.S. news text}” for political leaning, “\textit{Examine and evaluate the overall toxicity expressed in the given online text}” for toxicity, and “\textit{Examine and evaluate the overall sentiment expressed in the given movie review}” for sentiment. It then asks the model to return a JSON object containing a single final score on the requested scale.

\subsubsection{Rubric-Based Prompt}

The rubric-based prompt additionally includes a task-specific rubric that decomposes each construct (i.e., political leaning, sentiment, and toxicity) into conceptually relevant dimensions (see Appendix Table~\ref{tab:rubric_dims_compact}). The rubrics were tailored to each task and to each dataset’s unit of analysis, and grounded in the relevant literature. The model is asked to score every dimension on the requested scale and then to assign a single final score as a holistic judgment of the text as a whole. This design is intended to make the scoring process more structured by providing more detailed criteria for judgment. 

Under both prompting strategies, we analyze only the final score, so that the two prompts are compared on the same output. 

\subsection{Evaluation Framework}
\label{sec:evaluation}

We measure scale sensitivity by comparing the scores that the same model gives to the same text under two rating scale conditions that differ in exactly one design element. This section describes how we calculate numerical and semantic sensitivity.

\subsubsection{Normalization}

Because raw scores are marked on different scales, we first normalize all scores to a common $[0,1]$ range. Let $r_{i,s}$ denote the raw score assigned to item $i$ on scale $s$, and let $a_s$ and $b_s$ denote the scale's specified minimum and maximum values. The normalized score is
\begin{equation}
z_{i,s} = \frac{r_{i,s} - a_s}{b_s - a_s} \in [0,1].
\end{equation}

\subsubsection{Minimal Scale Pairs}

The 16 rating scale conditions cross response format (continuous, ordinal), scale range (3-, 5-, 11-, or 21-point) and scale sign (signed, unsigned).\footnote{Unsigned 3- and 5-point scales run from 1 to 3 and 1 to 5, whereas unsigned 11- and 21-point scales run from 0 to 10 and 0 to 20, ensuring that an exact midpoint exists in all cases.} To isolate sensitivity to each design element, we then compare only \textit{minimal pairs}, defined as two rating scale conditions that differ in one design element while the other two remain fixed. For example, comparing a continuous signed 5-point scale with an ordinal signed 5-point scale isolates sensitivity to response format because both scales have the same sign and range. This procedure leads to 40 minimal pairs. Eight pairs differ only in format, 24 differ only in range, and eight differ only in sign. We assess sensitivity to each design element using the pairs that differ in that element.

\subsubsection{Numerical Scale Sensitivity}

Numerical scale sensitivity measures how far a text's score moves when only the scale changes. For a minimal pair $(s,t)$ and the $N$ items with a valid score on both scales, it is the mean absolute difference between the normalized scores:
\begin{equation}
SS_{\mathrm{MAE}}(s,t) = \frac{1}{N} \sum_{i=1}^{N} \left| z_{i,s} - z_{i,t} \right|.
\end{equation}
$SS_{\mathrm{MAE}}$ ranges from 0 to 1. A value of 0.10, for example, means that scores move by one tenth of the full scale on average.

\subsubsection{Semantic Scale Sensitivity}

Semantic scale sensitivity measures whether a change in scale alters how the output is interpreted. To capture interpretation, we place each normalized score into one of three categories of equal width.
\begin{equation}
g(z) = \begin{cases}
\text{low}, & z < 1/3, \\
\text{middle}, & 1/3 \le z \le 2/3, \\
\text{high}, & z > 2/3.
\end{cases}
\end{equation}
These categories correspond to negative, neutral, and positive sentiment; to left, center, and right political leaning; and to content that is not toxic, moderately toxic, and extremely toxic. Semantic sensitivity is the share of items that fall into a different category on the two scales.
\begin{equation}
SS_{\mathrm{switch}}(s,t) = \frac{1}{N} \sum_{i=1}^{N} \mathbf{1}\!\left[\, g(z_{i,s}) \neq g(z_{i,t}) \,\right]
\end{equation}
$SS_{\mathrm{switch}}$ ranges from 0 to 1. For example, a value of 0.20 means that 20\% of texts change category when only one design element of the scale changes.

We use three categories because this is the finest distinction that all 16 rating scale conditions can express. The shortest scales offer only three values, low, neutral, and high, and every prompt defines the low end, the middle, and the high end of the scale, even when ordinal scales also label intermediate levels such as \textit{slightly} or \textit{strongly}. Splitting the normalized range into equal thirds places the three values of a scale with 3 points in three different categories and applies the same boundaries to every scale.

\subsection{Analytical Framework}

We calculate $SS_{\mathrm{MAE}}$ and $SS_{\mathrm{switch}}$ for each minimal pair within every combination of task, prompting strategy, and model. This yields 1{,}200 estimates per metric across 3 tasks, 2 prompting strategies, 5 models, and 40 minimal pairs. 

To answer RQ1, we average the estimates from the minimal pairs that change each design element, separately for format, range, and sign. We report these averages for numerical sensitivity, $SS_{\mathrm{MAE}}$ (RQ1a), and for semantic sensitivity, $SS_{\mathrm{switch}}$ (RQ1b). Each task, prompting strategy, model, and pair receives equal weight. We additionally analyze whether sensitivity to one design element depends on the values of the other two elements. For example, we calculate range sensitivity separately on ordinal and on continuous scales (i.e., format), and format sensitivity separately on scales with 3-, 5-, 11-, and 21-point scales (i.e., range).

To test RQ2 to RQ4, we fit a weighted least squares (WLS) regression of the 1{,}200 estimates on the design element changed (format, range, or sign), task, prompting strategy, and model. We include interactions between the design element changed and each of the other three factors. We fit the model separately for numerical sensitivity ($SS_{\mathrm{MAE}}$) and semantic sensitivity ($SS_{\mathrm{switch}}$).

We estimate uncertainty using an item bootstrap with 2{,}000 resamples. In each resample, we sample texts with replacement within each task and use the same sampled texts across all models, prompting strategies, and scale pairs. We then recompute every statistic and obtain 95\% confidence intervals from the 2.5th and 97.5th percentiles of the bootstrap estimates. 
Because every 95\% confidence interval lies within 0.01 of its estimate, we report point estimates in the Results section and give intervals in the figures and tables.

Lastly, we explore how much of the variation in sensitivity each factor explains to answer RQ5. We extend the regression to all interactions between pairs of design variables and decompose the variance of the 1{,}200 estimates into the share attributable to each factor, $\eta^2$. 


\section{Results}
\label{sec:results}

\subsection{Effects of Rating Scale Design}
\label{sec:rq1}

\begin{figure}[t]
    \centering
    \includegraphics[width=\columnwidth]{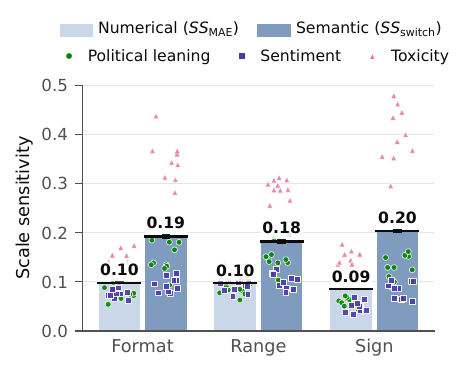}
    \caption{Scale sensitivity by rating scale design element. 
    Dots show the estimate for each combination of task, prompting strategy, and model, colored by task. Black lines are 95\% item bootstrap intervals.}
    \label{fig:factor_effects}
\end{figure}

RQ1 asks the extent to which changes in rating scale format (continuous vs.\ ordinal), range (3-, 5-, 11-, or 21-point), and sign (signed vs.\ unsigned) yield numerical and semantic sensitivity in LLM-generated outputs.

As shown in Figure~\ref{fig:factor_effects}, changing the \textbf{format} while holding sign and range fixed produces an average absolute difference of 0.098 in normalized scores (95\% CI [0.097, 0.099]) and changes the category assigned to 19.2\% of texts (95\% CI [18.9\%, 19.5\%]). Changing the \textbf{range} while holding format and sign fixed produces an average absolute difference of 0.098 in normalized scores (95\% CI [0.097, 0.099]) and changes the category assigned to 18.2\% of texts (95\% CI [17.9\%, 18.5\%]). Changing the \textbf{sign} while holding format and range fixed produces an average absolute difference of 0.085 in normalized scores (95\% CI [0.085, 0.086]) and changes the category assigned to 20.3\% of texts (95\% CI [20.1\%, 20.6\%]). 

Across all three design elements, numerical sensitivity amounts to approximately one tenth of the normalized scale, and semantic sensitivity indicates that one in five texts receives a different category when the scale design changes. The magnitudes of sensitivity are broadly similar across design elements. Specifically, changing the sign shows the smallest numerical sensitivity but the largest semantic sensitivity, which means a small move in a score can change its category when the score lies near a category boundary. 

\begin{figure}[t]
    \centering
    \includegraphics[width=\columnwidth]{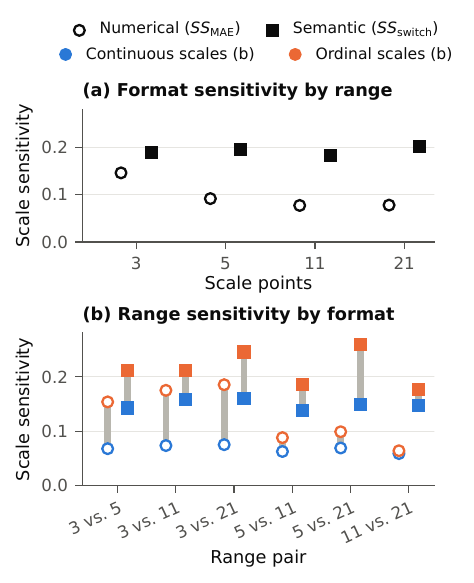}
    \caption{Format and range sensitivity by the settings of the other design elements. (a) Format sensitivity on scales with different rating ranges. (b) Range sensitivity for each pair of ranges. Bars are 95\% item bootstrap intervals.}
    \label{fig:scale_interactions}
\end{figure}

\begin{figure*}[t]
    \centering
    \includegraphics[width=\textwidth]{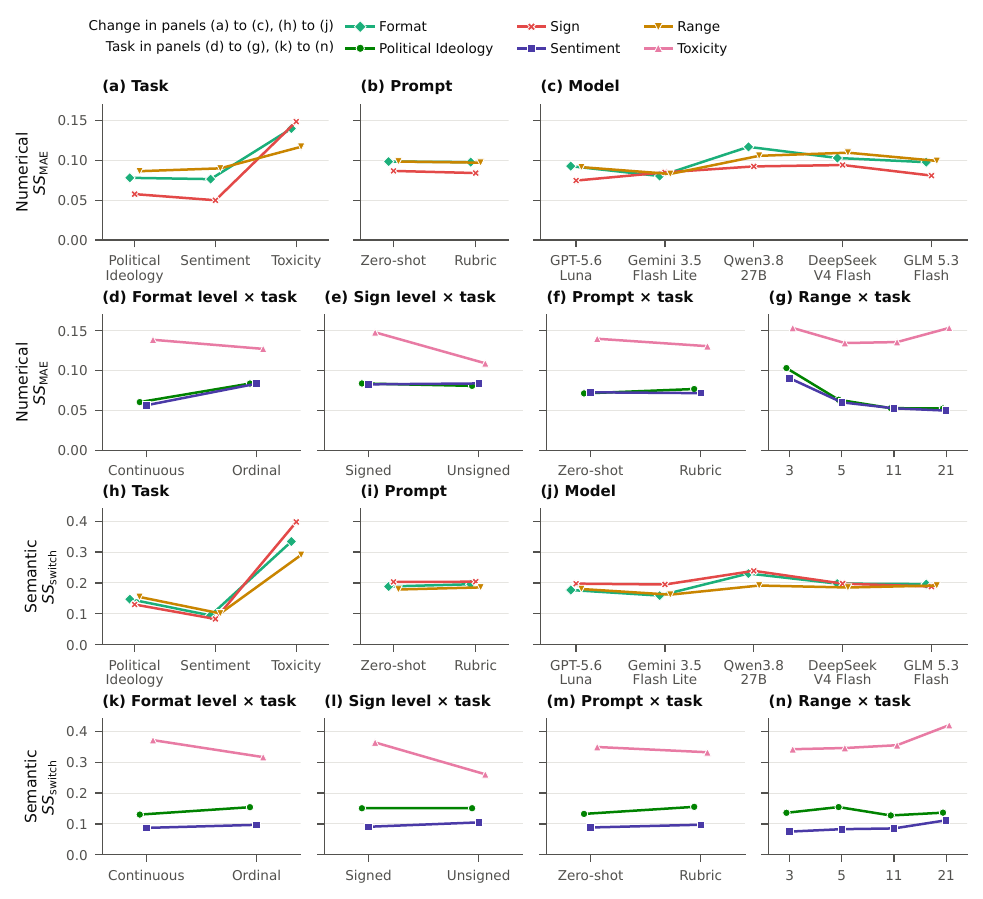}
    \caption{Scale sensitivity across tasks, prompting strategies, and models (RQ2--RQ4), averaged across the remaining tasks, prompting strategies, models, and minimal pairs. The top row shows numerical sensitivity, $SS_{\mathrm{MAE}}$, and the bottom row shows semantic sensitivity, $SS_{\mathrm{switch}}$. Error bars indicate 95\% item bootstrap confidence intervals.}
    \label{fig:moderation}
\end{figure*}

We additionally report whether sensitivity to one design element depends on the values of the other two elements, for example whether switching from a continuous to an ordinal format matters more on a 3-point scale than on a 21-point scale (see Figure~\ref{fig:scale_interactions}). We found three notable patterns. First, finer scales reduce numerical but not semantic sensitivity to \textbf{format}. Switching formats moves scores by 0.146 (95\% CI [0.144, 0.147]) on 3-point scales but by 0.078 ([0.077, 0.078]) on 21-point scales, whereas the share of texts that change category stays between 18.2\% and 20.1\% at every scale length. Second, ordinal scales amplify \textbf{range} changes, shifting scores by 0.128 ([0.126, 0.129]) compared with 0.068 ([0.067, 0.069]) on continuous scales and changing the category of 21.5\% compared with 14.9\% of texts. Third, continuous scales amplify \textbf{sign} changes, which alter the category of 24.4\% ([24.1\%, 24.7\%]) of texts on continuous scales but 16.3\% ([16.1\%, 16.6\%]) on ordinal scales. 

\subsection{Moderation by Task, Prompting Strategy, and Model}
\label{sec:rq2to4}


\begin{figure}[!t]
    \centering
    \includegraphics[width=\columnwidth]{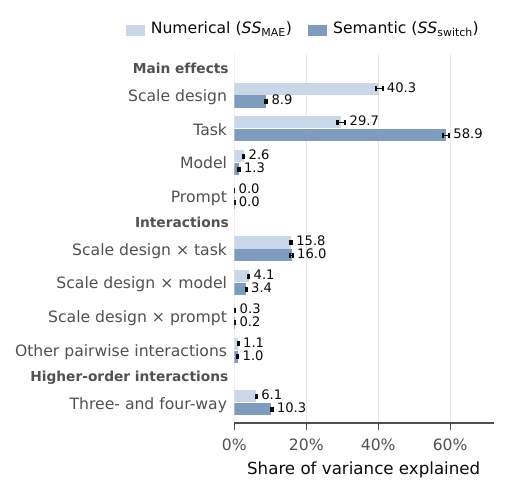}
    \caption{Share of the variation in scale sensitivity explained by each factor. 
    Black lines are 95\% item bootstrap intervals. The full decomposition is reported in Table~\ref{tab:variance_full} in Appendix~\ref{app:result_details}.}
    \label{fig:relative-importance}
\end{figure}

Concerning RQ2, scale sensitivity depends strongly on the task. Toxicity is the most sensitive, with 29\% to 40\% of texts changing category depending on the design element, compared with 13\% to 15\% for political leaning and 8\% to 10\% for sentiment. Numerical sensitivity follows the same order. The task also changes which design element produces the greatest sensitivity. For political leaning and sentiment, changes in the three design elements have similar effects, whereas for toxicity, changing the sign is the most sensitive. The gap between sign and format sensitivity is 8.0 percentage points larger for toxicity than for political leaning. Toxicity is the only unipolar construct in our design, ranging from absence to strong presence rather than between two opposing poles. This may explain why it reacts so strongly to signed scales, which place negative numbers at the low end of the scale, although with a single unipolar task we cannot rule out other differences between the tasks. Further details on the results are provided in Appendix~\ref{app:result_details}.

In contrast, the prompting strategy has almost no effect (RQ3). For every design element, sensitivity under rubric prompting is within 0.007 of sensitivity under zero-shot prompting, and all interaction terms are 0.006 or smaller (Table~\ref{tab:moderation}). The small differences that appear run in opposite directions across tasks. Rubrics slightly increase semantic sensitivity for political leaning and sentiment and slightly decrease it for toxicity. That is, adding explicit evaluation criteria does not make scores more stable to the rating scale.

For RQ4, model choice affects scale sensitivity more than prompting strategy but much less than task. All five models are sensitive to changes in all three design elements in the same way (see Figure~\ref{fig:moderation}). Qwen3.8 27B is the most sensitive model and Gemini 3.5 Flash Lite the least sensitive. Within each model, sensitivity to the three design elements differs by at most five percentage points. 

\subsection{Relative Importance of Each Design Choice}
\label{sec:rq5}

Regarding RQ5, the variance decomposition places all factors on a common scale (Figure~\ref{fig:relative-importance}) and shows that scale design and the task explain different aspects of sensitivity. Scale design explains the largest share of numerical sensitivity (40.3\%) but only 8.9\% of semantic sensitivity. The task explains 29.7\% of numerical sensitivity and 58.9\% of semantic sensitivity. 

The model and the prompting strategy explain little of either. The model accounts for 2.6\% of numerical and 1.3\% of semantic variation, and the prompting strategy for less than 0.1\%. Their interactions with scale design add 3.4\% to 4.1\% for the model and 0.2\% to 0.3\% for the prompting strategy. The second largest source of semantic variation is instead the interaction between scale design and task (16.0\%), which captures combinations such as signed scales for toxicity. 

\subsection{Additional Analyses}
\label{sec:add}

Moreover, we conduct two supplemental analyses. We further (a) test whether different levels of reasoning effort affect the results using the three open models (Appendix~\ref{app:reasoning}) and (b) assess the agreement between LLM-generated scores and the human labels provided in each dataset (Appendix~\ref{app:accuracy}). Notably, we find that higher reasoning effort is associated with small reductions in scale sensitivity. Agreement varies substantially across tasks and also across rating scale conditions within the same model and prompting strategy. Further details are provided in Appendix~\ref{additional-analyses}.

\section{Conclusion}
\label{conclusion}

This study investigates the extent to which LLM-generated scores remain consistent when only the rating scale design changes. Using an experimental design with 3 tasks, 2 prompting strategies, and 5 models, we find that changing scale format, range, or sign shifts normalized scores by about 0.10 on average and changes the category of roughly 20\% of texts. Toxicity, the only unipolar construct examined, is the most sensitive, particularly to scale sign. Rubric-based prompting does not reduce sensitivity relative to zero-shot prompting, and model choice has a modest effect.

Our findings indicate that rating scale design is a measurement decision that affects LLM-generated scores. We provide practical suggestions for future researchers.

First, we encourage researchers to test whether their main findings hold under at least one alternative rating scale. 
These robustness checks are essential for supporting the replicability and reliability of LLM-based measurement. 

Next, the numbers on the scale should match what the construct measures. For example, a unipolar construct such as toxicity ranges from absence to high intensity, so negative values on the rating scale may be difficult to interpret. In our experiments, models appeared to treat 0 as “not toxic” on signed scales, assigning substantially more texts to the high-toxicity category than on unsigned scales. Therefore, we recommend unsigned scales for unipolar constructs, with the lowest endpoint representing the absence of the attribute.

Also, researchers should consider evidence of validity when choosing a rating scale. When suitable human labels are available, they can compare scores across alternative scales with these labels to inform their choice. Ideally, the labels should come from trained annotators following coding guidelines aligned with the study’s goals. Agreement with these labels shows evidence of validity. When suitable reference labels are unavailable, as may often be the case for many social constructs, robustness checks across alternative scales become especially important for assessing the stability of the findings.

We acknowledge the following limitations. First, the generalizability of our findings is limited by the tasks and datasets examined. Future research is welcome to test whether the findings extend to broader cultural contexts, languages other than English, and additional tasks. 

Also, we compared only two prompting approaches: zero-shot and rubric-based. We did not include few-shot, chain-of-thought (CoT), or chain-of-verification (CoVe) prompting because these strategies introduce additional factors, such as the selection and ordering of demonstrations or intermediate reasoning steps, that could confound the comparison. Moreover, CoT has shown limited gains on subjective tasks \citep{chochlakis2024larger}, while CoVe is designed to verify factual claims \citep{dhuliawala2024chain}, making its applicability less clear for constructs that lack \textit{verifiable} ground truth. We leave the evaluation of these strategies to future work.

In short, careful scale design, robustness checks, and validation are needed to support more reliable use of LLM-generated scores. Our findings highlight the importance of these practices when using LLMs to complement human annotation in future studies.



\bibliography{custom}

\appendix
\raggedbottom
\section*{Appendix}

\section{Prompts}
\label{app:prompts}

This section outlines the prompts used for the scoring tasks, using political leaning as an example. The full prompts for all tasks and rating scale conditions are available at \url{https://anonymous.4open.science/r/Scale-Sensitivity-C418/}.

\subsection{Zero-Shot Prompt}
\label{app:prompt_formats}

Figure~\ref{fig:prompt-zeroshot} shows the zero-shot prompt for political leaning. The sentiment and toxicity prompts follow the same structure but use task-specific descriptions, text types (movie reviews for sentiment and online text for toxicity), and scale interpretation guidance under \texttt{\# Scale}. The toxicity prompt adjusts the scoring instructions to reflect its unipolar nature, assigning the lowest score when a text shows no toxicity and a middle score when toxicity is present but unclear. Greater distance from the lowest score indicates greater toxicity.

\begin{figure}[t]
\centering
\fbox{%
\begin{minipage}{0.95\linewidth}
\scriptsize\raggedright
\textbf{\# Task}\\
Examine and evaluate the overall political leaning expressed in the given U.S. news text.\\[0.3em]
\textbf{\# Scale}\\
\texttt{\{scale\_instructions\}}\\
- Use the requested scale for every score you output.\\
- For continuous scales, decimals are allowed.\\
- For discrete scales, output integer values only.\\
- Lower values mean more left-leaning. Higher values mean more right-leaning.\\
- For signed scales, negative values represent left-leaning, values near 0 represent neutral, mixed, balanced, or unclear leaning, and positive values represent right-leaning.\\
- For unsigned scales, lower values represent left-leaning, the midpoint or middle band represents neutral, mixed, balanced, or unclear leaning, and higher values represent right-leaning.\\[0.3em]
\textbf{\# Scoring Instructions}\\
- Base scoring only on evidence explicitly present in the given text.\\
- Do not infer political leaning from topic alone.\\
- If the evidence is absent, balanced, mixed, ambiguous, or unclear, use the neutral midpoint or neutral band of the requested scale.\\
- Use distance from the neutral midpoint or neutral band to reflect political leaning intensity.\\
- Use the full requested scale, including its endpoints, when the evidence supports it.\\
- Assign a single ``Final Political Leaning Score'' on the requested scale as a holistic judgment of the text as a whole.\\
- Output only valid JSON. Do not provide explanations, markdown, or extra keys.\\[0.3em]
\textbf{\# Output Format}\\
\texttt{\{"Final Political Leaning Score": \textless numeric value on the requested scale\textgreater\}}\\[0.3em]
\textbf{\# Text}\\
\texttt{\{text\}}
\end{minipage}}
\caption{Zero-shot prompt for political leaning.}
\label{fig:prompt-zeroshot}
\end{figure}

\subsection{Rubric-Based Prompt}
\label{app:rubric_dimensions}

The rubric-based prompt extends the zero-shot prompt with the additions shown in Figure~\ref{fig:prompt-rubric}. These include a \texttt{\# Marking Rubric} section after the scale section, three additional instructions under \texttt{\# Scoring Instructions}, and a separate output key for each rubric dimension. Table~\ref{tab:rubric_dims_compact} presents a condensed version of the rubric for each task.

For each task, we first defined the construct and its endpoints, then identified relevant dimensions in established research, codebooks, and annotation frameworks. 

\subsubsection{Political Leaning}
We drew on multidimensional accounts of ideology that distinguish economic and social attitudes \citep{TreierHillygus2009Ideology,FeldmanJohnston2014Ideology}. The Manifesto Project codebook \citep{Volkens2021ManifestoCodebook} provides policy categories covering economic intervention, welfare, traditional morality, civil liberties, law and order, and social groups. We adapted these categories to U.S. news text and organized them into economic policy, social and cultural values, government authority, and group-based equality. We also included explicit support for or criticism of political actors as partisan cues, yielding the five dimensions in Table~\ref{tab:rubric_dims_compact}.

\subsubsection{Sentiment}
We referred to distinctions between sentiment polarity, subjectivity, and degrees of evaluation \citep{PangLee2008OpinionMining}, along with VADER's treatment of sentiment intensity \citep{HuttoGilbert2014VADER}. We adapted these ideas to movie reviews through content evaluation, emotional tone, language intensity, and recommendation signals. The latter captures whether the reviewer encourages or discourages watching the film, complementing the review's evaluative language.

\subsubsection{Toxicity}
We adapted the attribute definitions in Jigsaw's Perspective API documentation \citep{PerspectiveAPIAttributes}. Hate and harassment combines criteria from toxicity and severe toxicity, while identity attack, insult, profanity, and threat correspond to the API's named attributes. We translated these definitions into evidence of increasing toxicity, with absence of the attribute at the low end.

\begin{figure}[t]
\centering
\fbox{%
\begin{minipage}{0.95\linewidth}
\scriptsize\raggedright
\textit{New section after \# Scale}\\
\textbf{\# Marking Rubric}\\
Score each dimension separately using this rubric:\\
| Dimension | What it measures | Left-Leaning Evidence | Neutral / Mixed / Unclear Evidence | Right-Leaning Evidence |\\
{[One row each for PC, EP, SC, GL, and IG (Table~\ref{tab:rubric_dims_compact}).]}\\[0.5em]
\textit{Added to \# Scoring Instructions}\\
- Not every text will address every dimension.\\
- Assign one numeric score to each dimension: PC, EP, SC, GL, and IG.\\
- The final score is not an arithmetic average of the dimension scores; weigh the dimensions by how much each one drives the overall political leaning of the text.\\[0.5em]
\textit{Added to \# Output Format, before the final score}\\
\texttt{"PC": \textless\ldots\textgreater, "EP": \textless\ldots\textgreater, "SC": \textless\ldots\textgreater, "GL": \textless\ldots\textgreater, "IG": \textless\ldots\textgreater,}
\end{minipage}}
\caption{Lines that the rubric-based prompt adds to the zero-shot prompt in Figure~\ref{fig:prompt-zeroshot}.}
\label{fig:prompt-rubric}
\end{figure}

\begin{table*}[t]
\centering\scriptsize
\setlength{\tabcolsep}{3pt}
\renewcommand{\arraystretch}{1.1}
\begin{tabular}{@{}p{0.09\textwidth}p{0.29\textwidth}p{0.30\textwidth}p{0.28\textwidth}@{}}
\toprule
\textbf{Task} & \textbf{Dimension and scope} & \textbf{Low-end evidence} & \textbf{High-end evidence} \\
\midrule
Political leaning
 & \textbf{PC} Party, candidate, movement, or political leaning. Political actors and groups.
 & Supports Democrats, liberals, or progressives, or criticizes conservatives from the left.
 & Supports Republicans, conservatives, MAGA, libertarians, or nationalists, or criticizes liberals from the right. \\
\addlinespace
 & \textbf{EP} Economic policy and redistribution. Economy, labor, taxation, markets, government intervention.
 & Supports redistribution, welfare, labor rights, progressive taxes, universal healthcare, or corporate regulation.
 & Supports lower taxes, deregulation, free markets, privatization, welfare cuts, limited government, or individual responsibility. \\
\addlinespace
 & \textbf{SC} Social and cultural values. Religion, immigration, reproductive and LGBTQ+ rights, diversity, secular governance.
 & Supports reproductive or LGBTQ+ rights, immigrant inclusion, multiculturalism, diversity, secularism, or progressive change.
 & Supports traditional or religious values, abortion restrictions, stricter immigration, assimilation, or opposition to progressive policies. \\
\addlinespace
 & \textbf{GL} Government authority, law, order, and security. State power, policing, justice, civil liberties.
 & Supports police accountability, justice reform, protest rights, civil liberties, rehabilitation, or limits on coercion.
 & Supports strong policing, punishment, enforcement, border control, military strength, surveillance, or public order. \\
\addlinespace
 & \textbf{IG} Identity, inclusion, and group-based equality. Discrimination, minority rights, systemic inequality, DEI.
 & Supports anti-discrimination, affirmative action, minority rights, systemic inequality claims, DEI, or marginalized groups.
 & Opposes affirmative action or DEI, or emphasizes meritocracy, colorblindness, or skepticism about systemic inequality. \\
\midrule
Sentiment
 & \textbf{Content Evaluation}. Overall evaluation of the film.
 & Film portrayed as poor, frustrating, disappointing, weak, or not worth watching.
 & Film portrayed as enjoyable, successful, valuable, exceptional, or highly satisfying. \\
\addlinespace
 & \textbf{Emotional Tone}. Reviewer's emotional attitude.
 & Angry, harsh, dismissive, sarcastic, hostile, disappointed, or strongly critical.
 & Warm, appreciative, satisfied, enthusiastic, admiring, or emotionally positive. \\
\addlinespace
 & \textbf{Language Intensity}. Strength and direction of evaluative language.
 & Negative evaluation, from mild criticism to strong condemnation.
 & Positive evaluation, from mild praise to strong enthusiasm. \\
\addlinespace
 & \textbf{Recommendation Signal}. Encouragement or discouragement to watch.
 & Discourages others from watching the film.
 & Recommends the film, possibly with strong encouragement. \\
\midrule
Toxicity
 & \textbf{HH} Hate and harassment. Hostile, degrading, disrespectful, or harassing language.
 & No highly hateful, aggressive, or extremely disrespectful language.
 & Very hateful, aggressive, degrading, or disrespectful language likely to discourage participation. \\
\addlinespace
 & \textbf{IA} Identity attack. Negative targeting based on identity.
 & No negative targeting of an identity group or characteristic.
 & Clearly hateful, demeaning, dehumanizing, or exclusionary targeting based on identity. \\
\addlinespace
 & \textbf{IN} Insult. Insults, mockery, or demeaning attacks.
 & No insult or demeaning language.
 & Clear insults, mockery, humiliation, or attacks on a person or group. \\
\addlinespace
 & \textbf{PR} Profanity. Swearing or obscene language, especially when abusive.
 & No profanity or obscene language.
 & Strong, repeated, or abusive profanity, especially directed at others. \\
\addlinespace
 & \textbf{TH} Threat. Intent, encouragement, or implication of harm or violence.
 & No suggestion of harm, violence, or intimidation.
 & Clear threats, or encouragement, celebration, or implication of violence or harm. \\
\bottomrule
\end{tabular}
\caption{Condensed marking rubrics. Each rubric in the original prompt also has a middle column for neutral, mixed, or unclear evidence (ambiguous or moderate evidence for toxicity), which we omit here. The full prompts are presented at \url{https://anonymous.4open.science/r/Scale-Sensitivity-C418/}}
\label{tab:rubric_dims_compact}
\end{table*}

\subsubsection{Scale Instructions}
\label{app:scale_instructions}

\texttt{\{scale\_instructions\}} is replaced by one sentence that states the format, range, and labels of the scale. For example, the unsigned 5-point ordinal scale for political leaning reads ``Rate the given text on a discrete ordinal scale from 1 to 5, where 1 indicates `far left', 2 indicates `left', 3 indicates `neutral', 4 indicates `right', and 5 indicates `far right'.'' Table~\ref{tab:ordinal_mappings} lists the labels for every task and number of points.

\begin{table*}[t]
\centering\scriptsize
\setlength{\tabcolsep}{3pt}
\renewcommand{\arraystretch}{1.1}
\begin{tabular}{@{}p{0.05\textwidth}p{0.29\textwidth}p{0.31\textwidth}p{0.31\textwidth}@{}}
\toprule
\textbf{Points} & \textbf{Political leaning} & \textbf{Sentiment} & \textbf{Toxicity} \\
\midrule
3  & left, neutral, right
   & negative, neutral, positive
   & not, moderately, extremely toxic \\
5  & far left, left, neutral, right, far right
   & very negative, negative, neutral, positive, very positive
   & not, slightly, moderately, strongly, extremely toxic \\
11 & far / moderate / slight left, neutral, slight / moderate / far right
   & extremely / moderately / slightly negative, neutral, slightly / moderately / extremely positive
   & not, slightly, somewhat, moderately, considerably, strongly, extremely toxic \\
21 & far / strong / moderate / slight left, neutral, slight / moderate / strong / far right
   & extremely / strongly / moderately / slightly negative, neutral, slightly / moderately / strongly / extremely positive
   & not, minimally, slightly, somewhat, moderately, considerably, strongly, severely, extremely toxic \\
\bottomrule
\end{tabular}
\caption{Labels in ordinal scale instructions, ordered from lowest to highest score.}
\label{tab:ordinal_mappings}
\end{table*}

\section{LLM Parameter Settings}
\label{app:llm_parameters}

This section details the generation settings of the five models.

As shown in Table~\ref{tab:llm_parameters}, for the three open models, we followed the sampling settings recommended on each model's Hugging Face page. The Qwen3.8 27B card recommends temperature 1.0, top-$p$ 0.95, top-$k$ 20, min-$p$ 0, presence penalty 0, and repetition penalty 1.0 for its thinking mode.\footnote{\url{https://huggingface.co/Qwen/Qwen3.8-27B}} The DeepSeek V4 Flash card recommends temperature 1.0 and top-$p$ 1.0 for tasks other than agentic use.\footnote{\url{https://huggingface.co/deepseek-ai/DeepSeek-V4-Flash-0731}} The GLM 5.3 Flash card gives no general recommendation, so we used temperature 1.0 and top-$p$ 0.95, the values the card reports for its own evaluations, and set \texttt{clear\_thinking} to true as the card advises for chat use.\footnote{\url{https://huggingface.co/zai-org/GLM-5.3-Flash}} Parameters that a card does not mention were left at the endpoint default. The main analyses use \texttt{low}, the lowest level of each open model, and Appendix~\ref{app:reasoning} reports the higher levels.

For the two proprietary models,  we ran GPT 5.6 Luna through the OpenAI Responses API and Gemini 3.5 Flash Lite through the Gemini API, using each provider's batch service. For both models we chose the lowest reasoning level that keeps reasoning enabled. For GPT 5.6 Luna this is \texttt{low}, since \texttt{none} turns reasoning off and the default is \texttt{medium}.\footnote{\url{https://developers.openai.com/api/docs/models/gpt-5.6-luna}} For Gemini 3.5 Flash Lite this is \texttt{minimal}, which is the model's default and which Google recommends for classification and extraction tasks.\footnote{\url{https://docs.cloud.google.com/gemini-enterprise-agent-platform/models/gemini/3-5-flash-lite}} We did not set temperature, top-$p$, or top-$k$ for either model. Google states that custom values for these parameters are not supported for Gemini 3.5 Flash Lite and are ignored, and OpenAI instructs users to omit temperature and top-$p$ when reasoning is enabled.\footnote{\url{https://developers.openai.com/api/docs/guides/latest-model}} 

\begin{table}[t]
\centering\scriptsize
\setlength{\tabcolsep}{5pt}
\begin{tabular}{@{}lllllr@{}}
\toprule
Model & Reasoning & Temp. & Top-$p$ & Top-$k$ \\
\midrule
GPT 5.6 Luna & low & default & default & default \\
Gemini 3.5 Flash Lite & minimal & default & default & default \\
Qwen3.8 27B & low & 1.0 & 0.95 & 20 \\
DeepSeek V4 Flash & low & 1.0 & 1.0 & default \\
GLM 5.3 Flash & low & 1.0 & 0.95 & default \\
\bottomrule
\end{tabular}
\caption{Generation settings for LLM scoring inference.}
\label{tab:llm_parameters}
\end{table}

\section{LLM Scoring: Output Description}
\label{app:output_description}

\begin{figure}[t]
    \centering
    \includegraphics[width=\columnwidth]{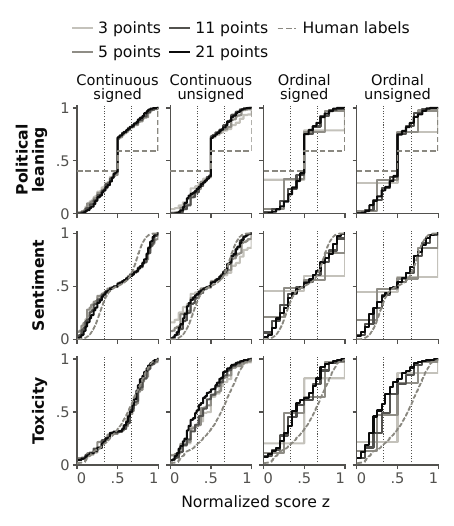}
    \caption{Cumulative distribution of normalized scores for each scale. Solid lines show the number of scale points, from light (3 points) to dark (21 points). The gray dashed line repeats the distribution of the human labels for that task in every panel. Dotted vertical lines mark the category boundaries at 1/3 and 2/3.}
    \label{fig:score_distributions}
\end{figure}

In this section, we describe the distributions of LLM-generated scores. Figure~\ref{fig:score_distributions} shows the cumulative distributions of normalized scores, with the human labels provided in each dataset shown as \textit{reference distributions}. Yet, we do not treat these labels as ground truth for the following reasons. First, the coding instructions followed by human annotators are not fully documented and may differ from our scoring criteria. Second, the political leaning labels are categorical, which limits direct comparison with our scalar scores. Third, this study focuses on the consistency of LLM-generated scores across rating scales rather than agreement between LLMs and human annotators. We include the human labels only to provide context for the observed score distributions.

\begin{figure}[t]
    \centering
    \includegraphics[width=\columnwidth]{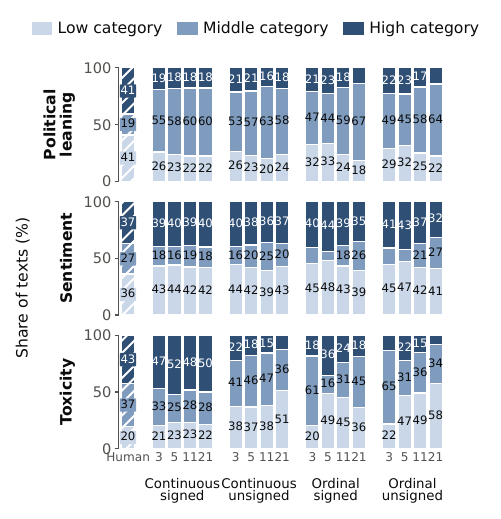}
    \caption{Share of texts in the low, middle, and high categories for each scale. Each row represents a task. 
    Percentages are displayed for segments exceeding 8\%.}
    \label{fig:category_composition}
\end{figure}

Figure~\ref{fig:category_composition} shows the proportion of texts categorized into low, middle, and high groups used to compute $SS_{\mathrm{switch}}$.

We observe distinct score distributions across tasks. Political leaning scores concentrate heavily in the middle category (44\%--67\% of texts, depending on the scale). Sentiment scores show a polarized distribution, with most texts falling into either the low (39\%--48\%) or high (32\%--44\%) category, and only 8\%--27\% in the middle. Toxicity scores vary highly depending on rating scales. On continuous signed scales, 47\%--52\% of texts fall in the high category, compared to only 12\%--22\% on continuous unsigned scales. This shift suggests that when a scale is centered on zero, models tend to treat 0—rather than the lowest endpoint—as the threshold for \textit{not toxic}.

When comparing with human labels, sentiment scores are closest to the human labels, although models use the middle category somewhat less often. Political leaning scores depart most clearly, as models place 44\% to 67\% of texts in the middle category compared with 19\% for the human labels. Toxicity categories vary most across scales, with the share of texts in the high category ranging from 8\% to 52\% compared with 43\% for the human labels. Model scores also concentrate on particular values such as the scale midpoint and whole numbers, which appears as vertical steps in Figure~\ref{fig:score_distributions}.

\section{Additional Analyses}
\label{additional-analyses}

\subsection{Reasoning Effort}
\label{app:reasoning}

The main analyses run every model at its lowest reasoning level. To test whether more reasoning reduces scale sensitivity, we repeat all 16 rating scale conditions with the three open models at three reasoning levels each, namely \texttt{low}, \texttt{medium}, and \texttt{xhigh} for Qwen3.8 27B, and \texttt{low}, \texttt{high}, and \texttt{max} for DeepSeek V4 Flash and GLM 5.3 Flash. Because the models name their levels differently, we compare them by their rank within each model (lowest, middle, highest). Sensitivity is calculated as in the main analyses. 

\begin{figure*}[t]
    \centering
    \includegraphics[width=\textwidth]{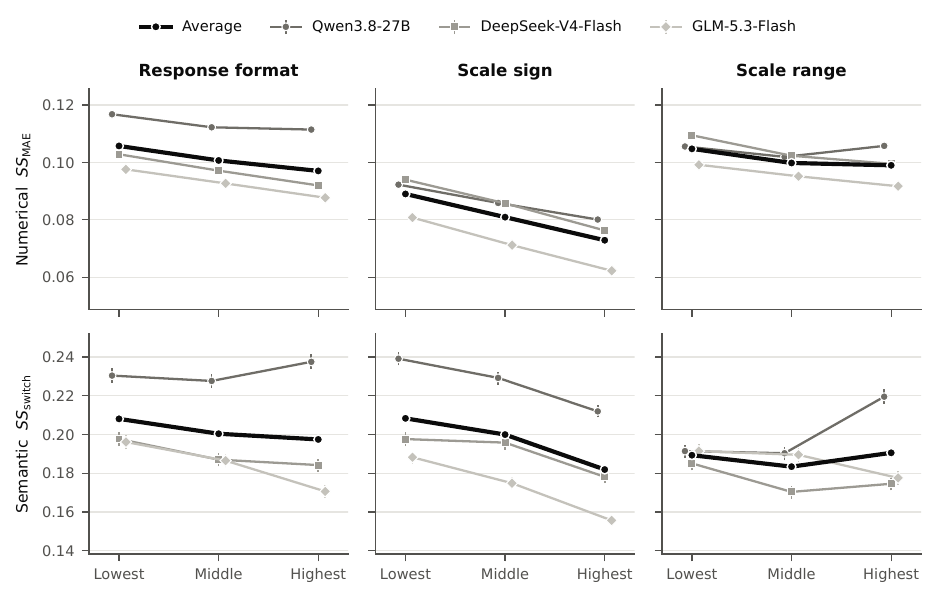}
    \caption{Scale sensitivity at three reasoning levels for the three open models. The black line is the average over the three models. Bars are 95\% item bootstrap intervals from 2{,}000 resamples.}
    \label{fig:reasoning}
\end{figure*}

Figure~\ref{fig:reasoning} shows the results. Interestingly, more reasoning lowers numerical sensitivity for changes in all three design elements, but only slightly. From the lowest to the highest level, $SS_{\mathrm{MAE}}$ falls by 0.009 for format, 0.006 for range, and 0.016 for sign, averaged over the three models. Semantic sensitivity falls by 1.1 percentage points for format and 2.6 points for sign, and does not change for range (the 95\% interval includes zero). The models also differ in direction. For Qwen3.8 27B, the share of texts that change category under a range change rises from 19.1\% to 22.0\%. Even at the highest level, sensitivity remains close to the values in Section~\ref{sec:results}, so more reasoning does not remove scale sensitivity.

\subsection{Agreement with Human Labels}
\label{app:accuracy}

\begin{figure*}[t]
    \centering
    \includegraphics[width=\textwidth]{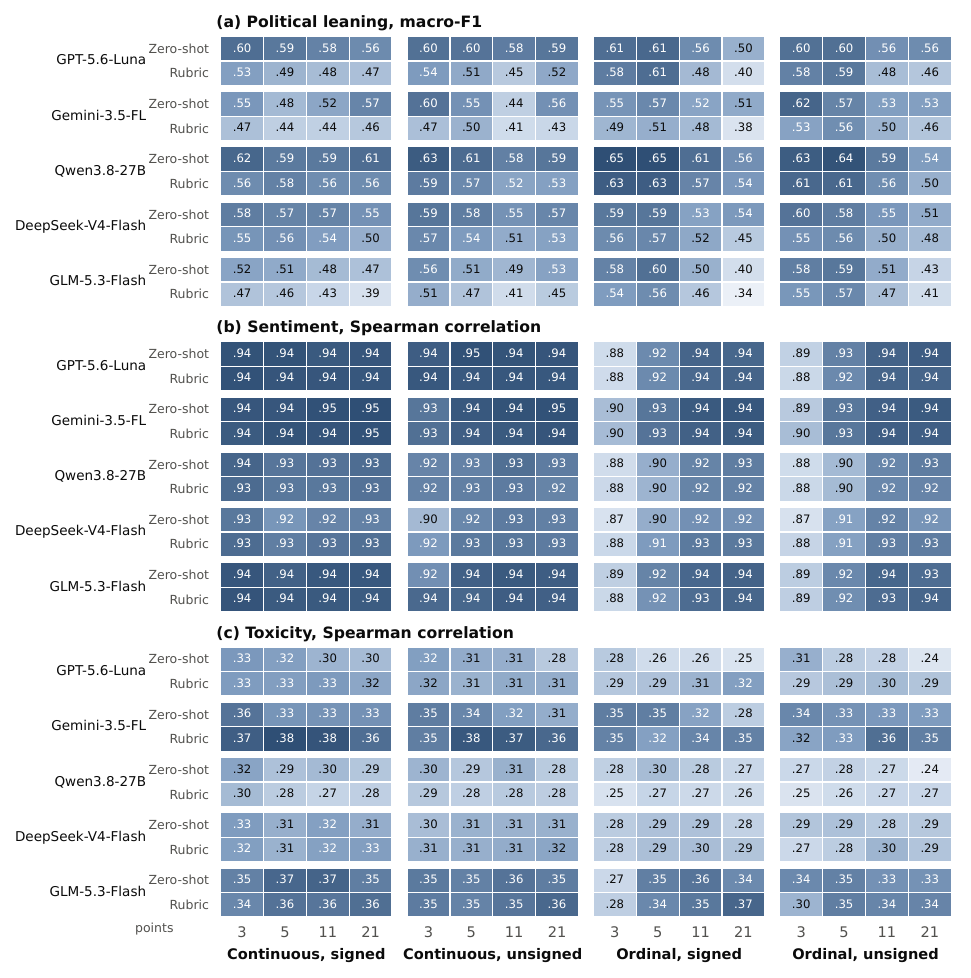}
    \caption{Agreement between LLM scores and human labels for every model, prompting strategy, and rating scale condition.}
    \label{fig:accuracy}
\end{figure*}

Our main analyses compare scores across rating scale conditions rather than against human labels. We additionally report how well the scores agree with the human labels of each dataset under every combination of model, prompting strategy, and rating scale condition (see Figure~\ref{fig:accuracy}). 

We select the agreement metric based on the type of human label. The sentiment labels in the VADER dataset \cite{HuttoGilbert2014VADER} and the toxicity labels in the Jigsaw Unintended Bias in Toxicity Classification dataset \cite{jigsaw-unintended-bias-in-toxicity-classification} are continuous. They represent the mean rating across annotators for sentiment and the proportion of annotators who marked a text as toxic for toxicity. For these tasks, we report Spearman's rank correlation between LLM-generated scores and human labels, which is unaffected by normalization to a common scale range. In contrast, the political leaning labels in MBIC \cite{Spinde2021MBIC} are categorical. They indicate whether the outlet that published each sentence is left, center, or right, based on media ratings rather than a judgment of the sentence itself. We map each LLM-generated score to the three categories used for $SS_{\mathrm{switch}}$ and report macro-F1 against the human labels.

Agreement with human labels varies substantially across tasks. Spearman's correlation ranges from 0.87 to 0.95 for sentiment and from 0.24 to 0.38 for toxicity, while macro-F1 ranges from 0.34 to 0.65 for political leaning. The lower correlations for toxicity may reflect what the human labels measure, namely the proportion of annotators who considered a text toxic rather than the intensity of its toxicity. Similarly, the lower agreement for political leaning may reflect that its labels describe the outlet rather than the sentence. MBIC also includes sentences on non-political topics, and a sentence from a left- or right-leaning outlet may express no leaning of its own.

Agreement also varies with rating scale design. For sentiment, 3-point ordinal scales yield the lowest correlations (0.87--0.90). For political leaning, average macro-F1 on ordinal scales decreases from 0.58 with 3 points to 0.48 with 21 points. Within each combination of model and prompting strategy, the difference between the highest and lowest macro-F1 across the 16 scales ranges from 0.09 to 0.23. Rubric-based prompting yields lower macro-F1 for political leaning than zero-shot prompting for every model and scale, averaging 0.51 compared with 0.56. For sentiment and toxicity, the average change in correlation between prompting strategies is less than 0.01.

\section{Result Details}
\label{app:result_details}

This section provides additional statistical details for the results presented in Section~\ref{sec:results}. All analyses were conducted in Python 3.9.6 with NumPy 2.0.2 and pandas 2.3.3. We fit the weighted least squares regressions with NumPy linear algebra routines and computed Spearman correlations with SciPy 1.13.1. Figures were created with Matplotlib 3.9.4.

First, Table~\ref{tab:moderation} reports the interaction coefficients from the weighted least squares (WLS) regression. Each sensitivity estimate is weighted by the inverse of the number of minimal pairs that vary the same design element, giving format, range, and sign equal total weight. Format is the reference level of the design element changed. Each interaction coefficient indicates how the difference in sensitivity between sign or range and format changes at a given moderator level compared with the moderator's reference level.

\begin{table*}[t]
\centering
\small
\setlength{\tabcolsep}{6pt}
\renewcommand{\arraystretch}{1.1}
\begin{tabular}{@{}llrlrl@{}}
\toprule
 & & \multicolumn{2}{c}{$SS_{\mathrm{MAE}}$} & \multicolumn{2}{c}{$SS_{\mathrm{switch}}$} \\
\cmidrule(lr){3-4}\cmidrule(l){5-6}
Moderator & Design element & $b$ & 95\% CI & $b$ & 95\% CI \\
\midrule
\multicolumn{6}{@{}l}{\textit{Task}} \\
Sentiment & Sign & $-0.006$ & [$-0.007$, $-0.005$] & $+0.006$ & [0.003, 0.008] \\
 & Range & $+0.005$ & [0.005, 0.006] & $-0.001$ & [$-0.003$, 0.001] \\
\addlinespace[3pt]
Toxicity & Sign & $+0.029$ & [0.028, 0.030] & $+0.080$ & [0.078, 0.084] \\
 & Range & $-0.031$ & [$-0.032$, $-0.030$] & $-0.051$ & [$-0.053$, $-0.049$] \\
\midrule
\multicolumn{6}{@{}l}{\textit{Prompting strategy}} \\
Rubric-based & Sign & $-0.002$ & [$-0.002$, $-0.001$] & $-0.006$ & [$-0.008$, $-0.005$] \\
 & Range & $-0.001$ & [$-0.001$, 0.000] & 0.000 & [$-0.002$, 0.001] \\
\midrule
\multicolumn{6}{@{}l}{\textit{Model}} \\
Gemini 3.5 Flash Lite & Sign & $+0.023$ & [0.022, 0.024] & $+0.015$ & [0.013, 0.018] \\
 & Range & $+0.004$ & [0.003, 0.005] & 0.000 & [$-0.002$, 0.002] \\
\addlinespace[3pt]
Qwen3.8 27B & Sign & $-0.006$ & [$-0.007$, $-0.006$] & $-0.012$ & [$-0.015$, $-0.009$] \\
 & Range & $-0.010$ & [$-0.011$, $-0.009$] & $-0.042$ & [$-0.045$, $-0.040$] \\
\addlinespace[3pt]
DeepSeek V4 Flash & Sign & $+0.009$ & [0.008, 0.010] & $-0.021$ & [$-0.023$, $-0.018$] \\
 & Range & $+0.008$ & [0.007, 0.009] & $-0.016$ & [$-0.018$, $-0.013$] \\
\addlinespace[3pt]
GLM 5.3 Flash & Sign & $+0.001$ & [0.000, 0.002] & $-0.028$ & [$-0.031$, $-0.025$] \\
 & Range & $+0.003$ & [0.002, 0.003] & $-0.008$ & [$-0.010$, $-0.006$] \\
\midrule
\multicolumn{2}{@{}l}{Weighted $R^2$} & \multicolumn{2}{l}{0.40} & \multicolumn{2}{l}{0.64} \\
\bottomrule
\end{tabular}
\caption{Moderation of scale sensitivity by task, prompting strategy, and model. Entries are interaction coefficients $b$ from weighted least squares regressions of the 1{,}200 minimal pair estimates, fitted separately for each metric. Brackets are 95\% item bootstrap intervals. Intervals that include zero indicate no detectable moderation.}
\label{tab:moderation}
\end{table*}

Next, Table~\ref{tab:variance_full} gives the full variance decomposition illustrated in Figure~\ref{fig:relative-importance}. In the variance decomposition, scale design distinguishes all 40 minimal pairs, so it covers which design element changes, the values compared for that element, and the values of the other two elements held fixed. Because every combination of minimal pair, task, prompting strategy, and model occurs exactly once, the shares do not overlap and add up to 100\%.

\begin{table}[!htbp]
\centering\scriptsize
\setlength{\tabcolsep}{2pt}
\begin{tabular}{@{}lrrr@{}}
\toprule
 & & \multicolumn{2}{c}{$\eta^2$ (\%)} \\
\cmidrule(l){3-4}
Component & df & $SS_{\mathrm{MAE}}$ & $SS_{\mathrm{switch}}$ \\
\midrule
\textbf{Scale design} & 39 & \textbf{40.3} & \textbf{8.9} \\
\quad Design element changed & 2 & 1.2 & 0.4 \\
\quad Format of compared scales & 2 & 13.6 & 4.6 \\
\quad Sign of compared scales & 2 & 3.2 & 1.6 \\
\quad Points or range pair & 11 & 17.0 & 1.6 \\
\quad Combinations of these settings & 22 & 5.4 & 0.7 \\
\textbf{Task} & 2 & \textbf{29.7} & \textbf{58.9} \\
\textbf{Model} & 4 & \textbf{2.6} & \textbf{1.3} \\
\textbf{Prompting strategy} & 1 & \textbf{0.0} & \textbf{0.0} \\
\midrule
\textbf{Scale design $\times$ task} & 78 & \textbf{15.8} & \textbf{16.0} \\
\quad Design element changed $\times$ task & 4 & 5.7 & 3.2 \\
\quad Scale settings $\times$ task & 74 & 10.1 & 12.8 \\
\textbf{Scale design $\times$ model} & 156 & \textbf{4.1} & \textbf{3.4} \\
\quad Design element changed $\times$ model & 8 & 0.7 & 0.5 \\
\quad Scale settings $\times$ model & 148 & 3.3 & 2.9 \\
\textbf{Scale design $\times$ prompting strategy} & 39 & \textbf{0.3} & \textbf{0.2} \\
\quad Design element changed $\times$ prompting strategy & 2 & 0.0 & 0.0 \\
\quad Scale settings $\times$ prompting strategy & 37 & 0.3 & 0.2 \\
\textbf{Other pairwise interactions} & 14 & \textbf{1.1} & \textbf{1.0} \\
\quad Task $\times$ model & 8 & 0.6 & 0.5 \\
\quad Task $\times$ prompting strategy & 2 & 0.3 & 0.4 \\
\quad Model $\times$ prompting strategy & 4 & 0.2 & 0.1 \\
\midrule
\textbf{Three- and four-way interactions} & 866 & \textbf{6.1} & \textbf{10.3} \\
\bottomrule
\end{tabular}
\caption{Full variance decomposition of the 1{,}200 minimal pair estimates (Figure~\ref{fig:relative-importance}). Shares come from the weighted least squares regression with all interactions between two factors and sum to 100\% for each metric. Subcomponents may not add exactly to their total because of rounding. All 95\% item bootstrap intervals lie within 1.2 percentage points of the estimates.}
\label{tab:variance_full}
\end{table}

\end{document}